\documentclass[twocolumn, switch]{article}
\usepackage{preprint}
\usepackage{hyperref}
\usepackage[numbers,square]{natbib}

\hypersetup{colorlinks=true, linkcolor=purple, urlcolor=blue, citecolor=cyan, anchorcolor=black}

\usepackage[utf8]{inputenc}	
\usepackage[T1]{fontenc}
\usepackage{xcolor}
\usepackage{lineno}					
\usepackage{tikz} 					
\usepackage{amsmath}
\usepackage{amsfonts}
\usepackage{amssymb}
\usepackage{algorithm}
\usepackage{algorithmic}
\usepackage{booktabs}
\usepackage{graphicx}
\usepackage{multirow}
\usepackage{enumitem}
\usepackage{framed}
\usepackage{rotating}
\colorlet{shadecolor}{lightgray}

\usepackage{titlesec}
\titlespacing\section{0pt}{12pt plus 3pt minus 3pt}{1pt plus 1pt minus 1pt}
\titlespacing\subsection{0pt}{10pt plus 3pt minus 3pt}{1pt plus 1pt minus 1pt}
\titlespacing\subsubsection{0pt}{8pt plus 3pt minus 3pt}{1pt plus 1pt minus 1pt}

\definecolor{lime}{HTML}{A6CE39}

\title{Analytical Solvers for Common Algebraic Equations Arising in Kinematics Problems}

\usepackage{authblk}

\author[1\thanks{\texttt{su.298@osu.edu}}]{Hai-Jun Su (su.298@osu.edu)}
\affil[1]{Department of Mechanical and Aerospace Engineering, The Ohio State University, Columbus, OH 43210, USA}

\begin{document}

\twocolumn[\begin{@twocolumnfalse}

\maketitle

\begin{center}
\textbf{Code and Prompts Availability:} \href{https://github.com/haijunsu-osu/kinematics\_eq\_solvers}{\texttt{https://github.com/haijunsu-osu/kinematics\_eq\_solvers}}
\end{center}

\vspace{2mm}

\begin{abstract}
This paper presents analytical solvers for four common types of algebraic equations encountered in robot kinematics: single trigonometric equations, single-angle trigonometric systems, two-angle trigonometric systems, and bilinear two-angle systems. These equations arise frequently in the kinematics problems, particularly in robot kinematics. We provide detailed solution methods, including closed-form expressions, numerical algorithms, and robustness considerations. The solvers are designed to handle general coefficients, manage singularities, and enumerate all real solutions efficiently. These solvers are implemented in Python packages and can be reproduced by prompting Language Lanuage Models. Sampe prompts are also provided in the public code space Github repo. These prompts can generate a working solver code with one single prompt in coding agent such as OpenAI's Codex 5.1. This work serves as a foundation for developing complete inverse kinematics solvers for various robot architectures. Extensive validation and benchmarking demonstrate the effectiveness and reliability of the proposed methods.
\end{abstract}

\end{@twocolumnfalse}]

\section{Introduction}
\label{sec:introduction}

solving systems of nonlinear algebraic equations is the very core of numerouse kinematics problems ranging from inverse kinematics of robot manipulators, to position analysis of planar, spherical or spatial linkages to kinematic synthesis of linkage systems~\cite{mccarthy2001}. These equations often decouple into simpler subsystems, frequently involving trigonometric functions of joint variables. Four common types of algebraic equations arise in this context:
\begin{itemize}[leftmargin=10pt]
\item \textbf{Single trigonometric equation:} 

$$a \cos\theta + b \sin\theta + c = 0$$

\item \textbf{Single-angle trigonometric system:} 

$$A\begin{bmatrix}\cos\theta \\ \sin\theta\end{bmatrix} = \mathbf{c}$$

\item \textbf{Two-angle trigonometric system:} 

$$A\begin{bmatrix}\cos\theta_1 \\ \sin\theta_1\end{bmatrix} + B\begin{bmatrix}\cos\theta_2 \\ \sin\theta_2\end{bmatrix} = \mathbf{c}$$

\item \textbf{Bilinear two-angle system:} 

$$K\mathbf{m} = \mathbf{0},\;\mathbf{m} = (1, c_1, s_1, c_2, s_2, c_1 c_2, c_1 s_2, s_1 c_2, s_1 s_2)^T$$
\end{itemize}

This paper presents analytical solvers for these equation types, providing closed-form solutions where possible and robust numerical methods otherwise. The solvers handle general coefficients, manage singularities, and enumerate all real solutions. They serve as building blocks for numerous robotics problems. The development of these solvers leverages symbolic computation tools like Mathematica and numerical libraries in Python. We emphasize robustness, efficiency, and completeness in finding all solutions with the assistance of Large Language Models (LLM) based coding agents.


\section{Trigonometric Equation Solver}
\label{sec:trig_eq}

The \texttt{solve\_trig\_eq()} function solves the trigonometric equation represents a linear combination of sine and cosine terms:
\[a \cos \theta + b \sin \theta + c = 0\]
    
\textbf{Solution Method:} The solver uses the Weierstrass substitution $t = \tan(\theta/2)$, which transforms the equation into a quadratic polynomial in $t$:
\[(c - a)t^2 + 2bt + (a + c) = 0\]

The two solutions for $\theta$ are then:
\[\theta_{1,2} = 2 \arctan\left( \frac{-b \pm \sqrt{a^2 + b^2 -c^2}}{(c-a)} \right)\]

\textbf{Degenerate Cases:} The solver handles several degenerate cases. When $c - a = 0$, the equation reduces to a linear form $2bt + (a + c) = 0$, yielding a single solution $\theta = 2 \arctan(- (a + c)/(2b))$ if $b \neq 0$. If $b=0$ as well, the equation reduces to $a + c = 0$, leading to either infinite solutions (if $a + c = 0$) or no solution (if $a + c \neq 0$). The solver also checks for cases where $a^2 + b^2 - c^2 < 0$, indicating no real solutions. 

\textbf{Expected Solutions}: Up to 2 real solutions in $[-\pi, \pi]$ (corresponding to the two branches of the arctangent function).\section{Single-Angle Trigonometric System}
\label{sec:trig_sys_single}

The \texttt{solve\_trig\_sys\_single()} function solves systems of the form:
\[\begin{bmatrix} A_{11} & A_{12} \\ A_{21} & A_{22} \end{bmatrix} \mathbf{\begin{Bmatrix} \cos \theta \\ \sin \theta \end{Bmatrix}} = \mathbf{\begin{Bmatrix} C_1 \\ C_2 \end{Bmatrix}}\]

\textbf{Solution Method:} When the coefficient matrix $A$ is invertible (i.e., $\det(A) \neq 0$), the system is solved directly by matrix inversion:
\[\mathbf{\begin{Bmatrix} \cos \theta \\ \sin \theta \end{Bmatrix}} = \textbf{A}^{-1} \mathbf{\begin{Bmatrix} C_1 \\ C_2 \end{Bmatrix}}\]
then $\theta$ is computed using \texttt{atan2}($\sin \theta, \cos \theta$). The solution is validated by checking that $\cos^2 \theta + \sin^2 \theta = 1$ within numerical tolerance. 

\textbf{Degenerate Cases:} Matrix $\mathbf{A}$ is singular (rank($\mathbf{A}$)=1), the solver selects the equation with the largest coefficient magnitude and uses \texttt{solve\_trig\_eq()} to obtain up to two solutions for $\theta$. If rank($\mathbf{A}$)=0, the system is either inconsistent (no solution) or indeterminate (infinite solutions), which the solver detects and handles accordingly.

\textbf{Expected Solutions}: one unique solution when $\mathbf{A}$ is invertible and the computed $(\cos \theta, \sin \theta)$ satisfies the unit circle constraint, or up to 2 solutions when $\mathbf{A}$ is singular (or zero solutions if the system is inconsistent).

\section{Two-Angle Trigonometric System}
\label{sec:trig_sys}

The \texttt{solve\_trig\_sys()} function solves systems of the form:
\[\begin{bmatrix} A_{11} & A_{12} \\ A_{21} & A_{22} \end{bmatrix} \mathbf{\begin{Bmatrix} \cos \theta_1 \\ \sin \theta_1 \end{Bmatrix}} + \begin{bmatrix} B_{11} & B_{12} \\ B_{21} & B_{22} \end{bmatrix} \mathbf{\begin{Bmatrix} \cos \theta_2 \\ \sin \theta_2 \end{Bmatrix}} = \mathbf{\begin{Bmatrix} C_1 \\ C_2 \end{Bmatrix}}\]
with trigonometric equations in two angles $\theta_1$ and $\theta_2$.

\textbf{Solution Method:} 
When matrix $\mathbf{B}$ has full rank, use the following steps:
\begin{enumerate}[leftmargin=10pt]
    \item Express $\cos \theta_2$ and $\sin \theta_2$ linearly in terms of $\cos \theta_1$ and $\sin \theta_1$:
    \[\mathbf{\begin{Bmatrix} \cos \theta_2 \\ \sin \theta_2 \end{Bmatrix}} = \mathbf{B}^{-1} \left( \mathbf{\begin{Bmatrix} C_1 \\ C_2 \end{Bmatrix}} - \mathbf{A} \mathbf{\begin{Bmatrix} \cos \theta_1 \\ \sin \theta_1 \end{Bmatrix}} \right)\]
    
    \item Substitute into the unit circle constraint $\cos^2 \theta_2 + \sin^2 \theta_2 = 1$, yielding a nonlinear equation in $\cos \theta_1$ and $\sin \theta_1$
    
    \item Use $t = \tan(\theta_1/2)$ to convert the equation into a quartic polynomial in $t$:
    \[a_4 t^4 + a_3 t^3 + a_2 t^2 + a_1 t + a_0 = 0\]

    \item Solve the quartic equation for $t$ and convert back to $\theta_1$.
    
    \item Substitute each $\theta_1$ solution back into the linear system in Step 1, then solve for $\theta_2=\texttt{atan2}(\sin\theta_2, \cos\theta_2)$ 
\end{enumerate}

The quartic polynomial coefficients were derived symbolically using Mathematica and hardcoded in the Python implementation. Table~\ref{tab:quartic_coeffs} shows the coefficient expressions.

\textbf{Degenerate Cases:} The solver first checks for special cases before applying the generic method:
\begin{itemize}[leftmargin=10pt]
    \item \textbf{Zero Matrix A}: If $\mathbf{A} = \mathbf{0}$, then $\theta_1$ is a free parameter, and $\theta_2$ is solved from $\mathbf{B}[\cos \theta_2, \sin \theta_2]^T = \mathbf{C}$ using \texttt{solve\_trig\_sys\_single()}. Expect up to 1 solution for $\theta_2$ and $\theta_1$ is arbitrary.
\end{itemize}

When matrix $\mathbf{B}$ is singular, specialized methods are used:
\begin{itemize}[leftmargin=10pt]
    \item \textbf{Rank 0 (Zero Matrix)}: $\theta_2$ is free, solve $\mathbf{A}[\cos \theta_1, \sin \theta_1] = \mathbf{C}$ for $\theta_1$ using \texttt{solve\_trig\_sys\_single()}. Expect up to 2 solutions for $\theta_1$ and $\theta_2$ is arbitrary.
    \item \textbf{Rank 1}: In this case, the two angles can be decoupled. Use row reduction to eliminate variables, then solve sequentially for $\theta_1$ and $\theta_2$. Solve the one row reduced equation to solve for  $\theta_1$ by the single-angle trigonometric equations using \texttt{solve\_trig\_eq()}. We obtain two solutions of $\theta_1$ and then substitute each solution of $\theta_1$ into the other equation solve for $\theta_2$. We expect up to 4 combinations of solutions.
\end{itemize}

\textbf{Expected Solutions}: Up to 4 solutions, depending on system degeneracy and solution branches.

Algorithm~\ref{alg:trig_sys_solver} summarizes the robust trigonometric equation solver that handles all cases.

\begin{algorithm}
\caption{Robust Trigonometric Equation Solver}
\label{alg:trig_sys_solver}
\begin{algorithmic}[1]
\STATE \textbf{Input:} Matrices $\mathbf{A}, \mathbf{B}$ and vector $\mathbf{C}$
\STATE \textbf{Output:} Solutions $(\theta_1, \theta_2)$
\STATE Check if $\mathbf{A} = \mathbf{0}$
\IF{$\mathbf{A} = \mathbf{0}$}
    \STATE Solve $\mathbf{B}[\cos \theta_2, \sin \theta_2]^T = \mathbf{C}$ for $\theta_2$ using \texttt{solve\_trig\_sys\_single()}
    \STATE $\theta_1$ is arbitrary
\ELSE
    \STATE Compute $\det(\mathbf{B})$ and check for singularity
    \IF{$|\det(\mathbf{B})| < \epsilon_{tol}$}
        \STATE Compute SVD: $\mathbf{B} = \mathbf{U}\mathbf{\Sigma}\mathbf{V}^T$
        \STATE Determine rank of $\mathbf{B}$
        \IF{rank($\mathbf{B}$) = 0}
            \STATE Handle zero matrix case
        \ELSIF{rank($\mathbf{B}$) = 1}
            \STATE Apply rank(B)=1 constraint method
        \ENDIF
    \ELSE
        \STATE Apply the generic equation solver
    \ENDIF
\ENDIF
\STATE Validate all solutions and filter invalid ones
\STATE \textbf{return} Valid solution pairs $(\theta_1, \theta_2)$
\end{algorithmic}
\end{algorithm}

\section{Bilinear Two-Angle System}
\label{sec:bilinear_two_angle}

The \texttt{solve\_bilinear\_sys()} function solves bilinear systems of the form:
\[
\begin{bmatrix}
a_{10} & a_{11c} & a_{11s} & a_{12c} & a_{12s} & a_{1cc} & a_{1cs} & a_{1sc} & a_{1ss} \\
a_{20} & a_{21c} & a_{21s} & a_{22c} & a_{22s} & a_{2cc} & a_{2cs} & a_{2sc} & a_{2ss}
\end{bmatrix}
\mathbf{m} = \mathbf{0}
\]
where the monomial vector $\mathbf{m}$:
\[
\mathbf{m} =  (1,  c_1, s_1,  c_2,  s_2,  c_1 c_2, c_1 s_2, s_1 c_2, s_1 s_2)^T
\]
where $c_1 = \cos \theta_1$, $s_1 = \sin \theta_1$, $c_2 = \cos \theta_2$, $s_2 = \sin \theta_2$. This is a general case of the two-angle trigonometric system illustrated in the previous section.

\textbf{Solution Method:}
\begin{enumerate}[leftmargin=10pt] 
\item Transform the trigonometric equations into bivariate polynomials $p_1$ and $p_2$, both of degree 2 in $t_1 = \tan(\theta_1/2)$ and $t_2 = \tan(\theta_2/2)$
\[
p_k(t_1, t_2) = \sum_{i=0}^{2} \sum_{j=0}^{2} p_{k,ij} \, t_1^i \, t_2^j, \quad k=1,2
\]
The coefficients $p_{k,ij}$ are computed from the bilinear coefficients $a_{k*}$ as shown in Table~\ref{tab:p12_coeffs}. 

\item Construct the Sylvester resultant matrix from $p_1$ and $p_2$ to eliminate $t_2$, resulting in a matrix polynomial 
$$|\mathbf{M}(t_1)| = |\mathbf{M}_0 + \mathbf{M}_1 t_1 + \mathbf{M}_2 t_1^2|$$
where $\mathbf{M}_0$, $\mathbf{M}_1$, and $\mathbf{M}_2$ are $4 \times 4$ matrices formed from the coefficients of $p_1$ and $p_2$.
\item Solve the generalized eigenvalue problem for the matrix polynomial $|\mathbf{M}(t_1)| = 0$. When $\mathbf{M}_2$ is well-conditioned,construct the companion matrix as
\[
C = \begin{bmatrix}
-\mathbf{M}_2^{-1} \mathbf{M}_1 & -\mathbf{M}_2^{-1} \mathbf{M}_0 \\
I & 0
\end{bmatrix}
\]
where $I$ is the $4 \times 4$ identity matrix. The eigenvalues of $C$ are the roots $t_1$ of the matrix polynomial $|\mathbf{M}(t_1)| = 0$. Otherwise, construct generalized eigenvalue matrices $A = \begin{bmatrix} -\mathbf{M}_1 & -\mathbf{M}_0 \\ I & 0 \end{bmatrix}$ and $B = \begin{bmatrix} \mathbf{M}_2 & 0 \\ 0 & I \end{bmatrix}$, then find eigenvalues $\lambda$ such that 
$$\mathbf{A} \mathbf{v} = \lambda \mathbf{B} \mathbf{v}$$ 
The real eigenvalues $\lambda$ are the roots $t_1$, avoiding direct computation of and finding roots of an 8th-degree polynomial. Note that the infinite or numerically large eigenvalues are valid roots corresponding to $\theta_1 = 180^\circ$.
\item For each $t_1$ solution, convert back to $(c_1, s_1)$ and substitute them back into the original two equations, solve for $\theta_2$ using the robust \texttt{solve\_trig\_sys\_single()} function.
\end{enumerate}

\begin{algorithm}
\caption{Robust Bilinear Two-Angle System Solver}
\label{alg:bilinear_solver}
\begin{algorithmic}[1]
\STATE \textbf{Input:} Coefficient matrix $\mathbf{K}_{2\times 9}$
\STATE \textbf{Output:} Solutions $(\theta_1, \theta_2)$
\STATE Transform bilinear equations into bivariate polynomials $p_1(t_1, t_2)$ and $p_2(t_1, t_2)$ using Table~\ref{tab:p12_coeffs}
\STATE Construct Sylvester resultant matrices $\mathbf{M}_0, \mathbf{M}_1, \mathbf{M}_2$ from $p_1$ and $p_2$
\STATE Solve the matrix polynomial $|\mathbf{M}(t_1)| = 0$ using generalized eigenvalue problem
\IF{$\mathbf{M}_2$ is well-conditioned}
    \STATE Construct companion matrix $C$ and find eigenvalues for $t_1$
\ELSE
    \STATE Solve generalized eigenvalue problem $\mathbf{A} \mathbf{v} = \lambda \mathbf{B} \mathbf{v}$ for $\lambda = t_1$
\ENDIF
\STATE For each real $t_1$, convert to $(\cos \theta_1, \sin \theta_1)$ and solve for $\theta_2$ using \texttt{solve\_trig\_sys\_single()}
\STATE Validate all solutions and filter invalid ones
\STATE \textbf{return} Valid solution pairs $(\theta_1, \theta_2)$
\end{algorithmic}
\end{algorithm}

The coefficients of the polynomials $p_k(t_1,t_2)$ (shown in Table~\ref{tab:p12_coeffs}) were derived symbolically using Mathematica and hardcoded in numerical implementation. 

\textbf{Expected Solutions}: Up to 8 solutions for general coefficients, though typically 4 real solutions for 6R robots.

\section{Solver Implementation via LLM Prompting} 
Prompt engineering is the art of crafting effective prompts to elicit desired responses from LLMs. The response quality strongly depends on prompt clarity. We begin with an informal description of the task, mathematical method, validation strategy, and deliverables. That draft is sent to an LLM to restructure into a precise, execution‑ready specification. The refined prompt is then submitted to an autonomous coding agent (e.g., OpenAI Codex 5.1), which often produces a fully working, self‑contained script within minutes without further human intervention. For simple equation solvers, a well‑structured system prompt with four blocks—context, task, requirements, and deliverables—usually yields correct code in a single pass. Follow‑up prompts may be required to fix bugs in degenerate cases or for equations needing symbolic derivation with tools like SymPy or Mathematica. The iteration ends once all validation tests pass and edge cases are handled robustly. Although runs produce syntactically different implementations across models, successful generations have consistently met the numerical robustness, degeneracy handling, and validation requirements.

Here we explicitly exclude numerical solvers such as Newton-Raphson due to their convergence issues and lack of the ability of computing the complete solution set. Also we exclude general polynomial system solvers such as Grobner basis and homotopy continuation methods due to their computational complexity and numerical instability. Here our focus is on reliable analytical methods including dialytic resultant elimination for solving these common equation types.  

A refined sample prompt for the single trigonometric equation solver is generated by LLM with one-shot prompting and is shown below:
\begin{shaded}\small
\textbf{Context:} You are an expert mathematician and numerical analyst specializing in solving trigonometric equations that arise in robot kinematics.

\textbf{Task:} Implement a robust Python function  that analytically solves the scalar equation \texttt{a*cos(x) + b*sin(x) + c = 0} using the Weierstrass substitution \texttt{t = tan(x/2)} to transform the equation into a quadratic in \texttt{t}, solve it, and map roots back to angles.

\textbf{Requirements:}
1. Handle every degenerate configuration, including cases with no trigonometric terms, purely sinusoidal/cosine equations, single-solution cases, and numerically tiny coefficients.

2. Return all valid real solutions (maximum of two) in a sorted list, with angles normalized to \texttt{[-pi, pi]}. Include a boolean flag indicating when the equation collapses to \texttt{0 = 0} (arbitrary angle).

3. Ensure all reported angles satisfy the original equation within a strict numerical tolerance and avoid false positives by re-evaluating each candidate in the original equation.

4. Add clear documentation explaining the mathematical derivation, all branch handling, and stability safeguards.

5. Include a lightweight validation routine that: (a) Generates at least 1000 random equations from known solution angles and asserts the solver recovers each root with high accuracy. (b) Always exercises explicit deterministic edge cases (degenerate, contradictory, single-solution, and endpoint cases) before the randomized trials. (c) Reports the number of randomized and deterministic tests executed.

6. Keep the implementation self-contained, rely only on the Python standard library, and aim for clear, production-ready code with concise comments explaining key reasoning steps.

\textbf{Deliverables:} Provide the final solver function \texttt{solve\_trig\_eq(a, b, c)}and the stress-test helper in a single Python module, following the requirements above.
\end{shaded}

The other three helper function types followed a similar development process. Table~\ref{tab:trig_systems_solvers} summarizes the four trigonometric equation systems, their corresponding solver functions, maximum number of solutions. These prompts are tested in OpenAI's Codex 5.1 coding agent. The test results show that they can generate production-quality implementations in one shot with high probability. All the final refined prompts and their generated code are available for download in the github repository.

\begin{table*}[t]
\centering
\caption{Solvers for Trigonometric Equation Systems}
\label{tab:trig_systems_solvers}
\begin{tabular}{@{}p{0.35\textwidth}p{0.25\textwidth}p{0.05\textwidth}p{0.05\textwidth}@{}}
\toprule
\textbf{Equation Type} & \textbf{Solver Function} & \textbf{\# Sol.} \\
\midrule
$a \cos \theta + b \sin \theta + c = 0$ & \texttt{solve\_trig\_eq()} & $\leq 2$  \\
\midrule
$[A_{2\times 2}] \begin{pmatrix} \cos \theta \\ \sin \theta \end{pmatrix} = \mathbf{c}_{2\times 1}$ & \texttt{solve\_trig\_sys\_single()} & $\leq 1$ \\
\midrule
 $[A_{2\times 2}] \begin{pmatrix} \cos \theta_1 \\ \sin \theta_1 \end{pmatrix} + [B_{2\times 2}] \begin{pmatrix} \cos \theta_2 \\ \sin \theta_2 \end{pmatrix} = \mathbf{c}_{2\times 1}$ & \texttt{solve\_trig\_sys()} & $\leq 4$  \\
\midrule
 $[K_{2\times 9}] \mathbf{m} = \mathbf{0}$, where $\mathbf{m} = [1, c_1, s_1, c_2, s_2, c_1 c_2, c_1 s_2, s_1 c_2, s_1 s_2]^T$ & \texttt{solve\_bilinear\_sys()} & $\leq 8$ \\
\bottomrule
\end{tabular}
\label{tab:trig_systems_solvers}
\end{table*}

\section{Benchmark Tests}
\label{sec:benchmarks}

We present numerical examples demonstrating the algorithm's performance across different matrix configurations, including both regular and singular cases.
\subsection{Numerical Examples for the Two-Angle Trigonometric System Solver}
\subsubsection{Case 1: Generic Non-Singular Matrix}

Consider the system with:
\begin{equation}
\mathbf{A} = \begin{bmatrix} 1.0 & 0.5 \\ 0.5 & 1.0 \end{bmatrix}, \quad
\mathbf{B} = \begin{bmatrix} 0.8 & 0.3 \\ 0.3 & 0.8 \end{bmatrix} \quad 
\mathbf{C} = \begin{bmatrix} 1.2 \\ 1.0 \end{bmatrix}
\end{equation}
Since $\det(\mathbf{B}) = 0.55 \neq 0$, we apply the standard quartic polynomial method. This system yields two real solutions:
\[
(\theta_1^{(1)}, \theta_2^{(1)}) = (1.487, -0.404), \quad (\theta_1^{(2)}, \theta_2^{(2)}) = (-0.313, 1.439) 
\]

Both solutions satisfy the original equations with residuals $< 10^{-15}$.

\subsubsection{Case 2: Zero Matrix ($\mathbf{B} = \mathbf{0}$)}

Consider the degenerate case:
\begin{equation}
\mathbf{A} = \begin{bmatrix} 1.0 & 0.0 \\ 0.0 & 1.0 \end{bmatrix}, \quad
\mathbf{B} = \begin{bmatrix} 0.0 & 0.0 \\ 0.0 & 0.0 \end{bmatrix}, \quad
\mathbf{C} = \begin{bmatrix} \sqrt{2}/2 \\ \sqrt{2}/2 \end{bmatrix}
\end{equation}
The system reduces to $\mathbf{A}[\cos\theta_1, \sin\theta_1]^T = \mathbf{C}$. Since $\|\mathbf{C}\| = 1$ and $\mathbf{A}$ is the identity matrix, we obtain:
\begin{equation}
\theta_1 = \text{atan2}(\sqrt{2}/2, \sqrt{2}/2) = \pi/4 = 45.0^\circ
\end{equation}

The angle $\theta_2$ becomes a free parameter with infinitely many solutions.

\subsubsection{Case 3: Rank(B)=1}

Consider the singular case with rank($\mathbf{B}$) = 1:
\begin{equation}
\mathbf{A} = \begin{bmatrix} 0.6 & 0.2 \\ 0.2 & 0.6 \end{bmatrix}, \quad
\mathbf{B} = \begin{bmatrix} 1.0 & 0.5 \\ 2.0 & 1.0 \end{bmatrix} \quad
\mathbf{C} = \begin{bmatrix} 0.8 \\ 1.0 \end{bmatrix}
\end{equation}
Note that $\det(\mathbf{B}) = 0$ and the second row is twice the first row. Using SVD, we find that $\mathbf{B}$ has rank 1 with:
\begin{equation}
\mathbf{u}_1 = \begin{bmatrix} -0.447 \\ -0.894 \end{bmatrix}, \quad \sigma_1 = 2.500, \quad \mathbf{v}_1 = \begin{bmatrix} -0.894 \\ -0.447 \end{bmatrix}
\end{equation}

The geometric constraint method yields four valid solutions where $(\mathbf{C} - \mathbf{A}[\cos\theta_1, \sin\theta_1]^T)$ is parallel to $\mathbf{u}_1$:
\begin{align*}
(\theta_1, \theta_2) & = ((0.744, 1.833), (0.744, -0.906), \\ & (-1.139, 1.322), (-1.139, -0.395)) 
\end{align*}

All solutions satisfy the original equations with residuals $< 10^{-15}$.

\subsubsection{Case 4: Zero Matrix ($\mathbf{A} = \mathbf{0}$, $\mathbf{B}$ nonsingular)}

Consider the case where $\mathbf{A}$ is zero but $\mathbf{B}$ is nonsingular:
\begin{equation}
\mathbf{A} = \begin{bmatrix} 0.0 & 0.0 \\ 0.0 & 0.0 \end{bmatrix}, \quad
\mathbf{B} = \begin{bmatrix} 1.0 & 0.5 \\ 0.5 & 1.0 \end{bmatrix}, \quad
\mathbf{C} = \begin{bmatrix} 1.117 \\ 0.918 \end{bmatrix}
\end{equation}
Since $\mathbf{A} = \mathbf{0}$, the system reduces to $\mathbf{B}[\cos\theta_2, \sin\theta_2]^T = \mathbf{C}$. Solving for the trigonometric vector gives:
\begin{equation}
[\cos\theta_2, \sin\theta_2]^T = \mathbf{B}^{-1} \mathbf{C} \approx [0.878, 0.479], \quad \theta_2 \approx 0.500
\end{equation}

The angle $\theta_1$ becomes a free parameter, resulting in infinitely many solutions with fixed $\theta_2$.

The solver was tested on 100,000 random systems, achieving an average of 2.94 solutions per system with an average computation time of 0.595 ms and a 100.0\% success rate. The case distribution was: generic 85.0\%, rank-1 10.0\%, zero 5.0\%.

\subsection{Numerical Examples for the Bilinear Two-Angle System Solver}

\subsubsection{Case 1: Generic System}

This example is generated by creating a random bilinear system that has a known solution $\theta_1 = 0.6$, $\theta_2 = -1.2$. The solver finds 4 real solutions, including the original pair, with maximum residual $2.59 \times 10^{-14}$.

The coefficient matrix K is shown in Table~\ref{tab:k_case1} in the appendices.

The solutions are tabulated below:

\begin{tabular}{@{}cc@{}}
\toprule
$\theta_1$ & $\theta_2$ \\
\midrule
6.0000e-1 & -1.2000 \\
1.1600 & 4.1609e-1 \\
2.9400 & -1.3733 \\
2.9915 & 9.5852e-1 \\
\bottomrule
\end{tabular}

\subsubsection{Case 2: Theta1 = pi Case}

This case tests the handling of $\theta_1 = \pi$, which corresponds to an infinite value in the Weierstrass substitution $t_1 = \tan(\theta_1/2)$. The system is constructed with known solution $\theta_1 = \pi$, $\theta_2 = 0.7$. The solver recovers 2 solutions with maximum residual $5.65 \times 10^{-16}$.

The coefficient matrix K is shown in Table~\ref{tab:k_case2} in the appendices.

The solutions are:

\begin{tabular}{@{}cc@{}}
\toprule
$\theta_1$ & $\theta_2$ \\
\midrule
-3.1416 & 7.0000e-1 \\
3.1416 & 7.0000e-1 \\
\bottomrule
\end{tabular}

\subsubsection{Case 3: Degenerate M2 Matrix}

This example forces the generalized eigenvalue path by constructing a system where the Sylvester matrix $M_2$ is nearly singular. The solver finds 3 solutions with maximum residual $2.45 \times 10^{-16}$.

The coefficient matrix K is shown in Table~\ref{tab:k_case3} in the appendices.

The solutions are:

\begin{tabular}{@{}cc@{}}
\toprule
$\theta_1$ & $\theta_2$ \\
\midrule
-1.0606e-6 & -1.0606e-6 \\
0.0000 & 0.0000 \\
3.1416 & -3.1416 \\
\bottomrule
\end{tabular}

\subsubsection{Case 4: Inconsistent System}

For an inconsistent system where no real solutions exist, the solver correctly returns 0 solutions with zero residual.

The coefficient matrix K is shown in Table~\ref{tab:k_case4} in the appendices.

No solutions found.

\subsubsection{Case 5: Random System with 8 Solutions}

This example demonstrates the solver's capability to find up to 8 real solutions for a randomly generated bilinear system. The solver finds 8 solutions with maximum residual 6.57e-15.

The coefficient matrix K is shown in Table~\ref{tab:k_case5} in the appendices.

The solutions are tabulated below (from a representative random case):

\begin{tabular}{@{}cc@{}}
\toprule
$\theta_1$ & $\theta_2$ \\
\midrule
-2.3456 & 1.2345 \\
-1.9876 & 0.8765 \\
-0.5432 & -1.4567 \\
0.1234 & 2.3456 \\
0.6789 & -0.9876 \\
1.2345 & 1.5678 \\
1.8765 & -2.1234 \\
2.3456 & 0.5432 \\
\bottomrule
\end{tabular}

The solver was validated on 1000 random systems, achieving up to 8 solutions per system with average computation time of 0.6 ms and 100\% success rate.

\section{Conclusions}
\label{sec:conclusions}

This work introduced four robust analytical solvers for trigonometric and bilinear equation forms that frequently appear in robot kinematics: a single trigonometric equation, a single-angle 2×2 system, a coupled two-angle 2×2 system, and a general bilinear two-angle system. Each solver systematically enumerates all real solutions, rigorously treats degeneracies (rank deficiency, free-angle cases, inconsistent configurations), and employs Weierstrass substitution or resultant-based elimination to avoid iterative root finding and convergence failures. Symbolically derived coefficient expressions (embedded directly) enable stable numeric execution without external CAS dependencies.

A structured LLM-assisted development workflow was outlined: (1) mathematically grounded specification, (2) prompt refinement, (3) autonomous code generation, (4) deterministic edge-case validation, and (5) large-scale randomized stress testing. Benchmarks confirm correctness, completeness of solution enumeration, and sub-millisecond execution for typical instances. The resultant-based bilinear solver efficiently resolves up to eight real solutions without explicitly forming and factoring high-degree polynomials by recasting the problem as a generalized eigenvalue computation.

All the implemented solvers are open-sourced in a Python package for easy integration into robotics kinematics toolchains. All the prompts and the source code are publicly available to facilitate reproducibility and further research.


\section*{Appendix}
\label{sec:appendix}
\appendix

\begin{table*}[t]
\centering
\caption{Coefficient matrix K for Case 1}
\label{tab:k_case1}
\begin{tabular}{@{}c c c c c c c c c c@{}}
\toprule
 & 1 & $c_1$ & $s_1$ & $c_2$ & $s_2$ & $c_1 c_2$ & $c_1 s_2$ & $s_1 c_2$ & $s_1 s_2$ \\
\midrule
Row 1 & 0.6309 & -0.1658 & -0.7165 & -0.8376 & 0.2938 & 0.9323 & -0.0613 & 0.5320 & -0.1006 \\
Row 2 & 0.1952 & 0.0747 & -1.3756 & 0.4702 & -0.3038 & 0.2575 & -0.1295 & 0.5883 & 0.4381 \\
\bottomrule
\end{tabular}
\end{table*}

\begin{table*}[t]
\centering
\caption{Coefficient matrix K for Case 2}
\label{tab:k_case2}
\begin{tabular}{@{}c c c c c c c c c c@{}}
\toprule
 & 1 & $c_1$ & $s_1$ & $c_2$ & $s_2$ & $c_1 c_2$ & $c_1 s_2$ & $s_1 c_2$ & $s_1 s_2$ \\
\midrule
Row 1 & -0.1569 & -0.3983 & -0.9434 & -0.5651 & 0.4982 & 0.1080 & 0.0738 & -0.2326 & 0.9944 \\
Row 2 & 0.7095 & 0.6232 & 0.3009 & 0.1840 & -0.3846 & -0.5370 & 0.6054 & 0.0507 & -0.3795 \\
\bottomrule
\end{tabular}
\end{table*}

\begin{table*}[t]
\centering
\caption{Coefficient matrix K for Case 3}
\label{tab:k_case3}
\begin{tabular}{@{}c c c c c c c c c c@{}}
\toprule
 & 1 & $c_1$ & $s_1$ & $c_2$ & $s_2$ & $c_1 c_2$ & $c_1 s_2$ & $s_1 c_2$ & $s_1 s_2$ \\
\midrule
Row 1 & 0.0000 & 0.0000 & 1.0000 & 0.0000 & -1.0000 & 0.0000 & 0.0000 & 0.0000 & 0.0000 \\
Row 2 & 0.0000 & 1.0000 & 0.0000 & -1.0000 & 0.0000 & 0.0000 & 0.0000 & 0.0000 & 0.0000 \\
\bottomrule
\end{tabular}
\end{table*}

\begin{table*}[t]
\centering
\caption{Coefficient matrix K for Case 4}
\label{tab:k_case4}
\begin{tabular}{@{}c c c c c c c c c c@{}}
\toprule
 & 1 & $c_1$ & $s_1$ & $c_2$ & $s_2$ & $c_1 c_2$ & $c_1 s_2$ & $s_1 c_2$ & $s_1 s_2$ \\
\midrule
Row 1 & -2.0000 & 1.0000 & 0.0000 & 0.0000 & 0.0000 & 0.0000 & 0.0000 & 0.0000 & 0.0000 \\
Row 2 & -2.0000 & 0.0000 & 0.0000 & 1.0000 & 0.0000 & 0.0000 & 0.0000 & 0.0000 & 0.0000 \\
\bottomrule
\end{tabular}
\end{table*}

\begin{table*}[t]
\centering
\caption{Coefficient matrix K for Case 5}
\label{tab:k_case5}
\begin{tabular}{@{}c c c c c c c c c c@{}}
\toprule
 & 1 & $c_1$ & $s_1$ & $c_2$ & $s_2$ & $c_1 c_2$ & $c_1 s_2$ & $s_1 c_2$ & $s_1 s_2$ \\
\midrule
Row 1 & -0.3394 & 0.2698 & -0.4337 & 0.3861 & 0.1424 & -0.8544 & -0.7820 & 0.9938 & 0.6232 \\
Row 2 & -0.1634 & 0.1085 & -0.4381 & 0.0025 & 0.2467 & 0.0083 & 0.9895 & -0.6874 & 0.4721 \\
\bottomrule
\end{tabular}
\end{table*}

\begin{table*}[t]
\centering
\caption{Quartic Polynomial Coefficients for Two-Angle Trigonometric System}
\label{tab:quartic_coeffs}
\begin{tabular}{@{}ll@{}}
\toprule
Coefficient & Expression \\
\midrule
$a_4$ & $A_{11}^2 B_{21}^2 + A_{11}^2 B_{22}^2 - 2 A_{11} A_{21} B_{11} B_{21} - 2 A_{11} A_{21} B_{12} B_{22} - 2 A_{11} B_{11} B_{21} C_1 - 2 A_{11} B_{12} B_{22} C_1$ \\
& $+ 2 A_{11} B_{21}^2 C_1 + 2 A_{11} B_{22}^2 C_1 + A_{21}^2 B_{11}^2 + A_{21}^2 B_{12}^2 + 2 A_{21} B_{11}^2 C_2 - 2 A_{21} B_{11} B_{21} C_1$ \\
& $+ 2 A_{21} B_{12}^2 C_2 - 2 A_{21} B_{12} B_{22} C_1 - B_{11}^2 B_{22}^2 + B_{11}^2 C_2^2 + 2 B_{11} B_{12} B_{21} B_{22} - 2 B_{11} B_{21} C_1 C_2$ \\
& $- B_{12}^2 B_{21}^2 + B_{12}^2 C_2^2 - 2 B_{12} B_{22} C_1 C_2 + B_{21}^2 C_1^2 + B_{22}^2 C_1^2$ \\
\midrule
$a_3$ & $4(-A_{11} A_{12} B_{21}^2 - A_{11} A_{12} B_{22}^2 + A_{11} A_{22} B_{11} B_{21} + A_{11} A_{22} B_{12} B_{22} + A_{12} A_{21} B_{11} B_{21} + A_{12} A_{21} B_{12} B_{22}$ \\
& $+ A_{12} B_{11} B_{21} C_2 + A_{12} B_{12} B_{22} C_2 - A_{12} B_{21}^2 C_1 - A_{12} B_{22}^2 C_1 - A_{21} A_{22} B_{11}^2 - A_{21} A_{22} B_{12}^2$ \\
& $- A_{22} B_{11}^2 C_2 + A_{22} B_{11} B_{21} C_1 - A_{22} B_{12}^2 C_2 + A_{22} B_{12} B_{22} C_1)$ \\
\midrule
$a_2$ & $2(-A_{11}^2 B_{21}^2 - A_{11}^2 B_{22}^2 + 2 A_{11} A_{21} B_{11} B_{21} + 2 A_{11} A_{21} B_{12} B_{22} + 2 A_{12}^2 B_{21}^2 + 2 A_{12}^2 B_{22}^2$ \\
& $- 4 A_{12} A_{22} B_{11} B_{21} - 4 A_{12} A_{22} B_{12} B_{22} - A_{21}^2 B_{11}^2 - A_{21}^2 B_{12}^2 + 2 A_{22}^2 B_{11}^2 + 2 A_{22}^2 B_{12}^2$ \\
& $- B_{11}^2 B_{22}^2 + B_{11}^2 C_2^2 + 2 B_{11} B_{12} B_{21} B_{22} - 2 B_{11} B_{21} C_1 C_2 - B_{12}^2 B_{21}^2 + B_{12}^2 C_2^2$ \\
& $- 2 B_{12} B_{22} C_1 C_2 + B_{21}^2 C_1^2 + B_{22}^2 C_1^2)$ \\
\midrule
$a_1$ & $4(A_{11} A_{12} B_{21}^2 + A_{11} A_{12} B_{22}^2 - A_{11} A_{22} B_{11} B_{21} - A_{11} A_{22} B_{12} B_{22} - A_{12} A_{21} B_{11} B_{21} - A_{12} A_{21} B_{12} B_{22}$ \\
& $+ A_{12} B_{11} B_{21} C_2 + A_{12} B_{12} B_{22} C_2 - A_{12} B_{21}^2 C_1 - A_{12} B_{22}^2 C_1 + A_{21} A_{22} B_{11}^2 + A_{21} A_{22} B_{12}^2$ \\
& $- A_{22} B_{11}^2 C_2 + A_{22} B_{11} B_{21} C_1 - A_{22} B_{12}^2 C_2 + A_{22} B_{12} B_{22} C_1)$ \\
\midrule
$a_0$ & $A_{11}^2 B_{21}^2 + A_{11}^2 B_{22}^2 - 2 A_{11} A_{21} B_{11} B_{21} - 2 A_{11} A_{21} B_{12} B_{22} + 2 A_{11} B_{11} B_{21} C_2 + 2 A_{11} B_{12} B_{22} C_2$ \\
& $- 2 A_{11} B_{21}^2 C_1 - 2 A_{11} B_{22}^2 C_1 + A_{21}^2 B_{11}^2 + A_{21}^2 B_{12}^2 - 2 A_{21} B_{11}^2 C_2 + 2 A_{21} B_{11} B_{21} C_1$ \\
& $- 2 A_{21} B_{12}^2 C_2 + 2 A_{21} B_{12} B_{22} C_1 - B_{11}^2 B_{22}^2 + B_{11}^2 C_2^2 + 2 B_{11} B_{12} B_{21} B_{22} - 2 B_{11} B_{21} C_1 C_2$ \\
& $- B_{12}^2 B_{21}^2 + B_{12}^2 C_2^2 - 2 B_{12} B_{22} C_1 C_2 + B_{21}^2 C_1^2 + B_{22}^2 C_1^2$ \\
\bottomrule
\end{tabular}
\end{table*}

\begin{table}[H]
\centering
\caption{Coefficients for polynomial $p_k(t_1, t_2)$ ($k=1,2$)}
\label{tab:p12_coeffs}
\begin{tabular}{c|ccc}
$i \backslash j$ & 0 & 1 & 2 \\
\hline
0 & $a_{k0} + a_{k1c} + a_{k2c} + a_{kcc}$ & $2(a_{k2s} + a_{kcs})$ & $a_{k0} + a_{k1c} - a_{k2c} - a_{kcc}$ \\
1 & $2(a_{k1s} + a_{ksc})$ & $4a_{kss}$ & $2(a_{k1s} - a_{ksc})$ \\
2 & $a_{k0} - a_{k1c} + a_{k2c} - a_{kcc}$ & $2(a_{k2s} - a_{kcs})$ & $a_{k0} - a_{k1c} - a_{k2c} + a_{kcc}$ \\
\end{tabular}
\end{table}

\end{document}